\documentclass[sigconf]{acmart}

\usepackage{threeparttable}
\usepackage{tabularx}
\usepackage{color}
\usepackage{booktabs}
\usepackage{verbatim}
\usepackage{amsopn}
\usepackage{amsmath}
\usepackage{multirow}
\usepackage{colortbl}
\usepackage{makecell} 

\usepackage{hyperref}
\hypersetup{
colorlinks=true,
linkcolor=red,
filecolor=magenta,      
urlcolor=blue,
}

\usepackage{amsfonts}
\usepackage{bbm}
\usepackage{bm}

\usepackage{xspace}

\makeatletter
\DeclareRobustCommand\onedot{\futurelet\@let@token\@onedot}
\def\@onedot{\ifx\@let@token.\else.\null\fi\xspace}

\def\eg{\emph{e.g}\onedot} 
\def\ie{\emph{i.e}\onedot} 
 
 \def\vs{\emph{vs}\onedot}

\makeatother
\usepackage{epstopdf}

\usepackage{bbding}
\usepackage{stmaryrd}

\AtBeginDocument{%
  }

\copyrightyear{2023} 
\acmYear{2023} 
\setcopyright{acmlicensed}\acmConference[MuSe '23]{Proceedings of the 4th Multimodal Sentiment Analysis Challenge and Workshop: Mimicked Emotions, Humour and Personalisation}{October 29, 2023}{Ottawa, ON, Canada}
\acmBooktitle{Proceedings of the 4th Multimodal Sentiment Analysis Challenge and Workshop: Mimicked Emotions, Humour and Personalisation (MuSe '23), October 29, 2023, Ottawa, ON, Canada}
\acmPrice{15.00}
\acmDOI{10.1145/3606039.3613105}
\acmISBN{979-8-4007-0270-9/23/10}

\begin{document}
\newsavebox\CBox
\def\textBF#1{\sbox\CBox{#1}\resizebox{\wd\CBox}{\ht\CBox}{\textbf{#1}}}
\title{
Exploiting Diverse Feature for Multimodal Sentiment Analysis
}


\author{Jia Li}
\authornote{Equally contribution}
\email{jiali@hfut.edu.cn}
\affiliation{%
    \institution{
    Hefei University of Technology}
    \city{Hefei}
    \state{Anhui}
    \country{China}
    \postcode{230601}
}
\author{Wei Qian}
\authornotemark[1]
\email{qianwei.hfut@gmail.com}
\affiliation{%
    \institution{
    Hefei University of Technology}
    \city{Hefei}
    \state{Anhui}
    \country{China}
    \postcode{230601}
}
\author{Kun Li}
\authornotemark[1]
\email{kunli.hfut@gmail.com}
\affiliation{%
    \institution{
    Hefei University of Technology}
    \city{Hefei}
    \state{Anhui}
    \country{China}
    \postcode{230601}
}
\author{Qi Li}
\authornotemark[1]
\email{liqiahucs@gmailc.om}
\affiliation{%
    \institution{
    Anhui University}
    \city{Hefei}
    \state{Anhui}
    \country{China}
    \postcode{230000}
}
\author{Dan Guo}
\email{guodan@hfut.edu.cn}
\authornote{Corresponding authors}
\affiliation{%
    \institution{
    Hefei University of Technology \\ Institute of Artificial Intelligence,
Hefei Comprehensive National
Science Center}
    \city{Hefei}
    \state{Anhui}
    \country{China}
    \postcode{230601}
}
\author{Meng Wang}
\authornotemark[2]
\email{eric.mengwang@gmail.com}
\affiliation{%
    \institution{
    Hefei University of Technology \\ Institute of Artificial Intelligence,
Hefei Comprehensive National
Science Center}
    \city{Hefei}
    \state{Anhui}
    \country{China}
    \postcode{230601}
}

\renewcommand{\shortauthors}{Anonymous et al.}

\begin{abstract}
In this paper, we present our solution to the MuSe-Personalisation sub-challenge in the MuSe 2023 Multimodal Sentiment Analysis Challenge. 
The task of MuSe-Personalisation aims to predict the continuous arousal and valence values of a participant based on their audio-visual, language, and physiological signal modalities data.
Considering different people have personal characteristics, the main challenge of this task is how to build robustness feature presentation for sentiment prediction. 
To address this issue, we propose exploiting diverse features. 
Specifically, we proposed a series of feature extraction methods to build a robust representation and model ensemble. 
We empirically evaluate the performance of the utilized method on the officially provided dataset. 
\textbf{As a result, we achieved 3rd place in the MuSe-Personalisation sub-challenge.}
Specifically, we achieve the results of 0.8492 and 0.8439 for MuSe-Personalisation in terms of arousal and valence CCC. 
\end{abstract}




\maketitle

\section{Introduction} 
\begin{figure}[t]
\centering
\includegraphics[width=1.0\linewidth]{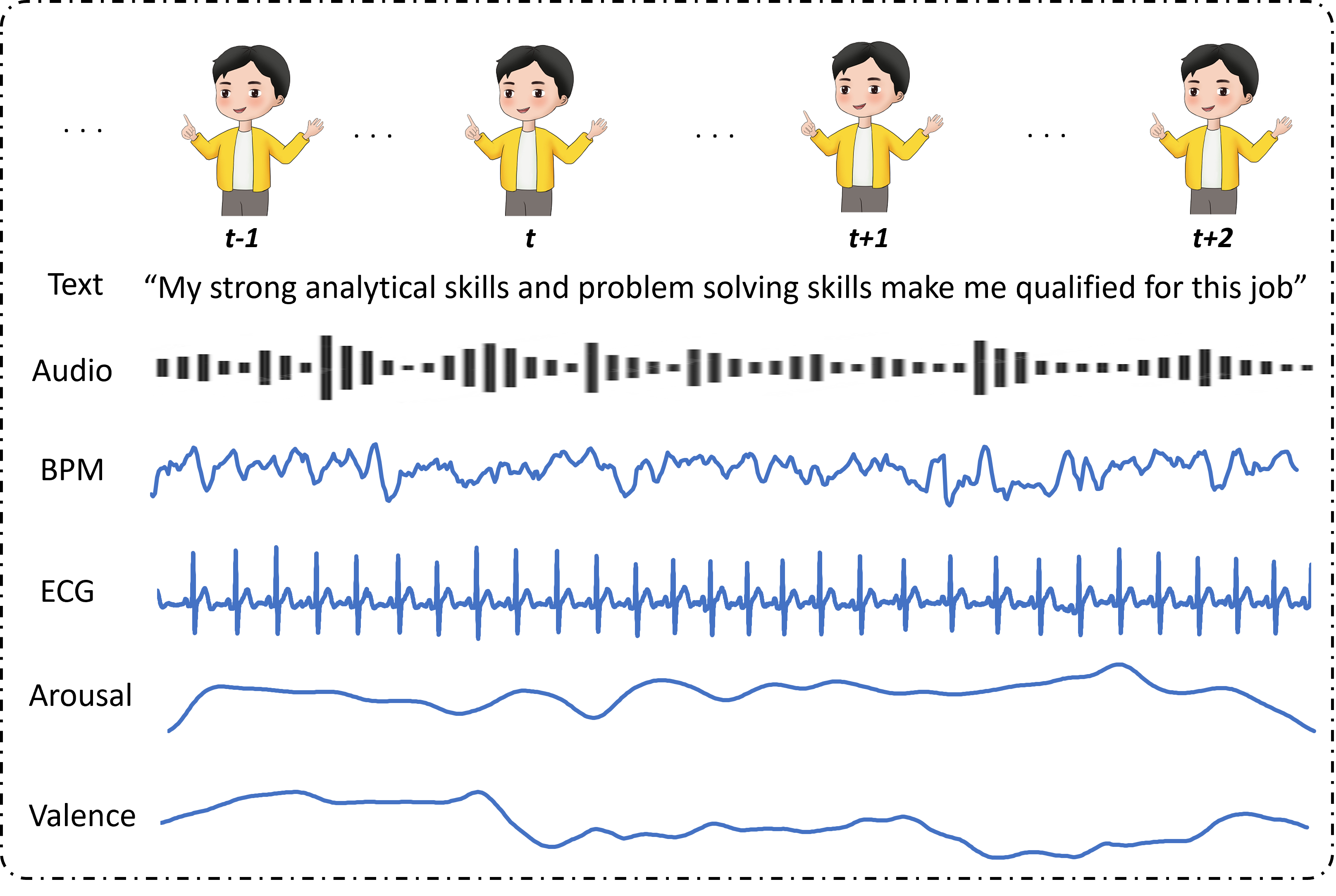}
\vspace{-1.0em}
\caption{Video sample from the Ulm-TSST dataset~\cite{stappen2021muse}. There are multiple modality features, such as visual, audio, and BPM. The task of MuSe-Personalisation aims to predict the continuous values of arousal and valence. }
\vspace{-1.5em}
\label{fig:feature}
\end{figure}
Sentiment analysis is one of the most important research issues in the research field of pattern recognition and has drawn extensive attention in recent years~\cite{xue2021transfer,xue2022vision,li2023multimodal,li2022hybrid}. The expression of sentiment is distributed in different modalities, such as facial expressions, words, speech, and physiological signals. Therefore, emotional analysis is naturally a task that relies on multimodal data. 

MuSe-Personalisation~\cite{christ2023muse} is a track of the 4$^{th}$ Multimodal Sentiment Analysis Challenge and Workshop at ACM MM 2023. 
It is based on the Ulm-TSST dataset introduced in MuSe 2021~\cite{stappen2021muse} and requires predicting the continuous arousal and valence values of a participant based on their audio-visual, language, and physiological signal modalities data. The Ulm-TSST dataset is an instance of the TSST dataset which defines a stress-inducing, job interview-like scenario. 
The initial state of the Ulm-TSST dataset consists of 110 individuals (about 10 hours), with rich annotations based on self-reported and continuous dimensional ratings of emotions (valence and arousal). 
Here, valence represents the level of emotional positivity, while arousal represents the level of excitement. 

To address the above issue, we leverage different methods to extract different modal features. We argue that more robust features can capture clues related to emotions and more accurately predict emotions.
It is necessary to predict the values of valence and arousal in a continuous manner. There is the problem of overfitting in both valence and arousal, especially the serious overfitting of arousal. There is a significant difference between the results of the development set and the test set, making it difficult to address this issue. 

In summary, our main contributions are summarized as follows:
\begin{itemize}
\item We extract new visual features and physiological features, leading to CCC improvement on the baseline model. In addition, we find that pre-trained visual features on emotional datasets are more robust. 
\item We explore different modality features effects \sloppy on Muse-Personalisation and found that some features play a negative role. In addition, we also explore different feature combinations to mine features and achieved 0.8439 on the combined CCC score. 
\item For model training, we explore the effects of RNN layers, the dimension of model, and whether RNN is bidirectional on performance in the baseline model. We found the baseline model is sensitive to the hidden dimension of features and the layer of RNN.  

\end{itemize}

The remainder of this paper is organized as follows. 
Related works are introduced in Section~\ref{sec:real}. 
In Section~\ref{sec:method}, we present the details of the utilized method and loss optimization. Section~\ref{sec:experiment} presents the implementation details and experiments to evaluate the performance the utilized model. 
The conclusion of our work is concluded in Section~\ref{sec:conclusion}.

\section{Related Work}\label{sec:real}
Various solutions have been proposed on the Ulm-TSST dataset~\cite{stappen2021muse} for MuSe Competition~\cite{christ2023muse,christ2022muse,amiriparian2022muse,stappen2021muse}. 
The Personalisation sub-challenge (MuSe-Personalisation) involves the regression task of physiological arousal and valence continuous signals. MuSe 2023 provides handcrafted features and deep features for each modality, which are made available for participants. The handcrafted features include audio features like eGeMAPS~\cite{valstar2016avec} provided by AVEC2016 and commonly used FAU~\cite{ekman1978facial} features in emotion recognition. However, with the advancement of neural network models and the increase in dataset size, deep features can offer richer information compared to handcrafted features. 

Multimodal data fusion strategies~\cite{li2021proposal,guo2018hierarchical,li2023vigt} can be categorized into early fusion, intermediate fusion, and late fusion. Most existing multimodal emotion recognition methods are based on early fusion, where the multimodal data is combined as a whole before performing the learning task. Joint representations can be directly extracted from vectors connected to deep models such as 1D-CNN~\cite{santamaria2018using} and Bi-LSTM~\cite{zitouni2021arousal}, which allow encoding the correlations between modalities. Late fusion methods~\cite{ayata2020emotion} integrate decisions from multiple independently learned models to predict emotion categories. Therefore, compared to early fusion, this fusion method ignores the relationships and interactions between modalities. For MuSe-Personalisation, we adopt an early fusion approach to integrate multimodal features.

\section{Method}\label{sec:method}
\begin{figure*}[t]
\centering
\includegraphics[width=1.0\linewidth]{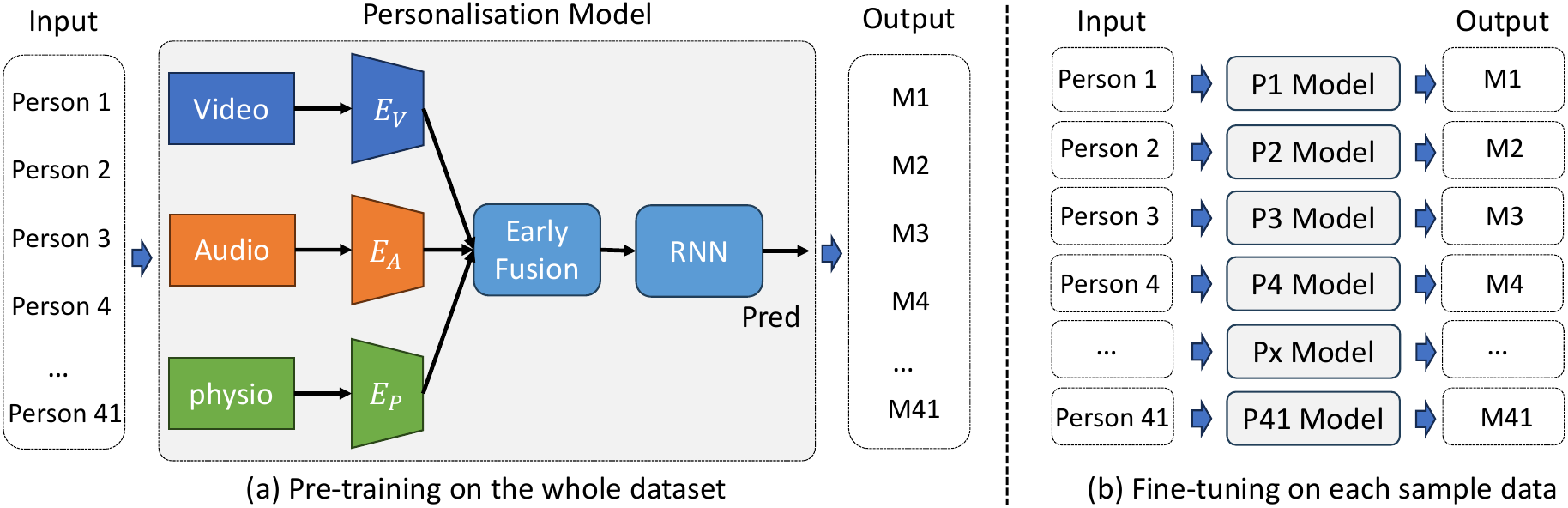}
\vspace{-2.0em}
\caption{The architecture overview of the baseline method in the MuSe-Personalisation sub-challenge. The model is trained in two steps: (a) the model is pre-trained on the whole dataset; (b) the model is then fine-tuned on each sample data.}
\vspace{-1.0em}
\label{fig:model}
\end{figure*}

Our method mainly consists of three parts, including {\bf feature preparation}, {\bf model architecture}, and {\bf loss function}.
\subsection{Feature Preparation}
Excellent feature preparation is necessary for good performance in emotion recognition. In this sub-section, we explore different audio-video features for multimodal emotion recognition.

\subsubsection{\bf Acoustic Feature}
DeepSpectrum~\cite{amiriparian2017snore} is a deep learning-based feature extraction method from audio signals using pre-trained image Convolutional Neural Networks (CNNs). The backbone of CNN chooses DenseNet121~\cite{huang2017densely} pre-trained on ImageNet. The obtained features are the output of the last pooling layers, which is a 1024-dimensional feature vector. 
WAV2VEC2.0~\cite{baevski2020wav2vec} is a self-supervised pretrained Transformer model in computer audition. The feature is extracted by averaging over the outputs in the final layer of this model, obtaining 1024-dimensional embeddings.
EGEMAPS is extracted by using the openSMILE toolkit~\cite{eyben2010toolkit}, which is an 88-dimensional extended Geneva Minimalistic Acoustic Parameter Set feature vector for sentiment analysis.
The above audio features are provided by the organizers of MuSe 2023.

\subsubsection{\bf Visual Feature}
In addition to the deep features such as FaceNet512~\cite{schroff2015facenet} and ViT~\cite{caron2021vit} introduced by MuSe 2023, we also introduce two additional deep features, namely emotion features extracted using the PosterV2-Vit~\cite{mao2023poster} model pretrained on the AffectNet~\cite{mollahosseini2017affectnet} dataset, and features obtained using the APVIT~\cite{apvit} model pretrained on the RAF-DB~\cite{li2017reliable} dataset.
AffectNet is built by using the pre-trained model PosterV2-Vit on the AffectNet dataset, resulting in a feature vector of 1536 dimensions. The PosterV2-Vit model was obtained from the joint training of the PosterV2~\cite{mao2023poster} and ViT models after individual pre-training on the AffectNet dataset. The final model achieved an accuracy of 67.57\% on the AffectNet test set. 
In order to extract comprehensive emotion-related facial features, we make use of the pre-trained APViT~\cite{apvit} method to process the extracted facial images and output a 767-
dimensional embedding for each image. APViT is a simple and efficient Transformer-based method for facial expression recognition. We take the released version, which is pre-trained on real-world facial expression recognition dataset RAF-DB~\cite{li2017reliable}.

\subsubsection{\bf physiological Feature}
Regarding the physiological signals, MuSe 2023 provides raw Electrocardiogram (ECG) signals with a sampling rate of 1000Hz, and the downsampled biosignals to 2 Hz by Savitzky-Golay filtering for smoothing. The Respiration rate (RESP) and heart rate (BPM) at a sampling rate of 2Hz are also given.   In addition to these three physiological signals, we utilized the NeuroKit2~\cite{makowski2021neurokit2} toolbox to extract 18 time-domain features related to heart rate variability (HRV) from the raw ECG data, such as MeanNN, SDNN, RMSSD, SDSD, and CVNN. Specifically, we design a sliding window size of 4000 steps (\ie, 4s) and divide the raw ECG signal with a hop size of 500 steps (\ie, 0.5s). The head and tail of the raw data are padded with neighboring data. Then we utilize the toolbox to extract the 18 time-domain features from each sliding window. Finally, by combining the three physiological signals provided by MuSe 2023, we obtained a 21-dimensional physiological feature representation, namely \textbf{Phys}.

\subsection{Model Architecture}
In this section, we depict the proposed model architecture in Figure~\ref{fig:model} and describe it in the following. 

\subsubsection{\bf Personalisation Model}
As shown in Figure~\ref{fig:model} (a), we use the multimodal feature (\eg, video, audio, and physiological signal) as the input of the personalisation model for physiological arousal or valence signal estimation.
The multimodal features are first extracted from the original audio-video data according to the above feature engineering methods. 
Then, we fuse the multimodal features by concatenating the audio, video, and physiological features along with the feature dimension. 
Specifically, we first simply concatenate the audio feature (\eg, DeepSpectrum), the video feature (\eg, Affectnet), and the physiological feature (\eg, Phys), the concatenated feature then is passed through a fully connected layer with weight $ W^{N \times d}$ to fuse the information of different modalities. We denote the unimodal feature by $ f \in \mathbb{R}^{T \times n_{i}} $, where $ i $ denotes the $i$th modality, $T$ and $n_{i}$ are sequence length and feature dimension, respectively. The process of early fusion is formulated as Eq. (~\ref{eq:cat}):
\begin{equation}\label{eq:cat}
F_{early} = Concat(f_{a} , f_{v}, f_{b})*W+b,
\end{equation}
where $f_{a}$, $f_{v}$, and $f_{b}$ denote the audio, visual, and physiological features, respectively. $N=n_1 + n_2 + n_3$ is the total dimension of concatenated feature and $d$ is set to 128 or 256 in experiments.

Considering the sequential data, we choose the Recurrent Neural Network (RNN) model as the backbone following the baseline paper.
More specifically, we adopt the Gate Recurrent Unit (GRU) to learn the
temporal information for predicting continuous estimations of physiological arousal and valence signals. The fused multimodal feature $F_{early}$ first passes through GRU layers to learn the sentimental feature. Then, the sentimental feature outputted by GRU layers is fed into two cascaded fully connected layers to regress the physiological arousal or valence signal. The above process can be formulated as:
\begin{equation}
H = GRU(F_{early}), \\
M = fc(\sigma(fc(H)),
\end{equation}
where $H$ and $M$ are the hidden semantic feature and emotion prediction. $\sigma$ and $fc$ denote the ReLU activation function and fully connected layer, respectively.

\subsubsection{\bf Fine-tuning}
Since the total number of MuSe-Personalisation sub-challenge is only 69 subjects, the proposed model needs to overcome the over-fitting problem. Following the strategy in the MuSe 2023 baseline paper, we perform fine-tuning in each test sample. Concretely, as shown in Figure~\ref{fig:model}(b), we duplicate the pre-trained model for every test subject and further train the model on the subjects’ data only. Note that, to find a better model for each subject, we set 10 fixed random seeds for training.
Finally, we choose the best model among the 10 subject-specific models for every test subject and predict the corresponding labels.

\subsubsection{Loss Function}
According to the baseline paper of Muse-Challenge, we take the Concordance Correlation Coefficient (CCC) as the loss function to train the proposed model. The CCC loss evaluates the tendency between predicted and target signals by scaling their correlation coefficient with their mean square difference. The value of CCC is in the range [-1, 1], and the higher the better. Therefore, the loss function can be formulated as follows: 
\begin{equation}
Loss =  1 - \frac{2\rho \delta_{x}\delta_{y}}{\delta_{x}^{2}+\delta_{y}^{2}+(\mu_{x}-\mu_{y})^2},
\end{equation}
where $\mu_{x}$ and $\mu_{y}$ denote the mean of the prediction $x$ and the label $y$, respectively. $\delta_{x}$ and $\delta_{y}$ are their standard deviations. $\rho$ indicates the Pearson correlation coefficient (PCC) between $x$ and $y$.

\section{Experiment}\label{sec:experiment}
\subsection{Dataset} 
The Personalisation Sub-Challenge (MuSe-Personalisation) aims to predict the continuous physiological arousal and valence values of participants based on their speech and video data. MuSe-2023 Challenge continues to use the Ulm-Trier Social Stress Test (Ulm-TSST) dataset introduced by MuSe-2021. As shown in Figure~\ref{fig:feature}, each subject must record approximately 5 minutes of video, including continuous images, speech, and textual data. In addition, the physiological signals, such as Electrocardiogram (ECG), Respiration rate (RESP), and heart rate (BPM), are recorded simultaneously at a sampling rate of 2Hz. The labels consist of two continuous emotion dimensions, \ie, physiological arousal and valence, obtained by combining the ratings of three annotators using the Rater Aligned Annotation Weighting (RAAW)~\cite{stappen2021muse} method, with a sampling rate of 2Hz.

\begin{table}[t]
\centering
\caption{Data statistics of the Ulm-TSST dataset at the MuSe-Personalisation challenge.}
\vspace{-1.0em}
\begin{tabular}{c|c|c|c}
\toprule
Partition    & \#Videos   & Duration &  \#Subjects  \\ \hline
Train & 41 (+14)   & 3:39:56  &  41 (+14)\\
Development  & 14 (+14)   & 1:20:10  &  41 (+14)\\
Test  & 14  & 0:47:21  &  41 (+14)\\ \bottomrule
\end{tabular}
\vspace{-1.5em}
\label{table:t1}
\end{table}

Unlike the commonly used subject-independent setting in previous works, the MuSe-Personalisation competition introduces partial open data in the test set to encourage the exploration of the adaptability of multimodal emotion recognition models to individuals. As shown in Table~\ref{table:t1}, the Ulm-TSST dataset includes a total of 69 participants with ages ranging from 18 to 39 years. The dataset is divided into training, development, and test sets based on the participants. The training set consists of 41 videos, and both the development and test sets contain 14 videos. Furthermore, to promote the personalized fitting of models to the test set, the first 60 seconds of each test subject's data is treated as additional subject-specific training data, the next 60 seconds as other subject-specific development data, and the remaining portion is used as the competition test set, which is not publicly available.

\begin{table*}[t]
\caption{CCC performance comparison of different features on the development and test set. A, V, T, and P denote audio, video, text, and physiological modality, respectively. Dimension denotes the dimension size of the feature.}
\vspace{-1.0em}
\begin{tabular}{c|c|c|cc|cc|c}
\toprule
\multirow{2}{*}{Modality } & \multirow{2}{*}{Feature} & \multirow{2}{*}{Dimension} & \multicolumn{2}{c|}{Arousal$\uparrow$} & \multicolumn{2}{c|}{Valence$\uparrow$} & Combined$\uparrow$ \\ \cline{4-8}
  &  &  & Dev.  & Test  & Dev.  & Test  & Test     \\ \hline
A & eGeMaps      & 88 & 0.7878 & 0.5523 & 0.6242 & 0.5472 & 0.5498   \\
A & DeepSpectrum  & 1024   & 0.8010 & 0.7447 & 0.5757 & 0.5185 & 0.6316   \\
A & Wave2Vec2.0   & 1024   & 0.7582 & 0.5053 & 0.4953 & 0.5117 & 0.5085   \\ \hline
V & FAU    & 20 & 0.6088 & 0.3476 & 0.4287 & 0.4653 & 0.4065   \\
V & ViT    & 384   & 0.3794 & 0.2480 & 0.6083 & 0.4805 & 0.3642   \\
V & FaceNet512    & 512   & 0.6413 & 0.5912 & 0.3337 & 0.3358 & 0.4635   \\
V & AffectNet    & 768   & 0.5527 & 0.6249 & 0.4779 & 0.5856 & 0.6052   \\
V & APViT  & 768   & 0.7568 & 0.5413 & 0.7937 & 0.7196 & 0.6304   \\ \hline
P & Phys   & 21   & 0.4028 & 0.2269 & 0.5758 & 0.4110 & 0.3189   \\ \hline
T & BERT   & 768 & 0.3245 & 0.1613 & 0.5174 & 0.3102 & 0.2358   \\ \bottomrule
\end{tabular}
\vspace{-0.5em}
\label{table:t2}
\end{table*}

\subsection{Experimental Setup}
{\bf Implementation Details.} We follow the baseline setting, applying a window size of 200 steps (100 s) and a hop size of 100 steps (50 s) to the first-step training of valence, a window size of 10 steps (5 s) and a hop size of 5 steps (2.5 s) are used in the second-step training for valence. For physiological arousal, we find that changing the window size in the first-step training to 200 steps (100 seconds) and the hop size to 100 steps (50 seconds) leads to better performance. The parameter setting of the second step training is consistent with valence evaluation. 

For model training, we employ the early fusion model for physiological arousal and valence evaluation. Specifically, we concatenate unimodal features in the sequence dimension and feed it into the RNN or Bi-RNN with 256 dimensions in the hidden state. According to our experience, the number of RNN layers is set to 1 which can obtain the best performance and generalization. Our early fusion models are trained with Adam optimizer and the initial learning rate is 1e-3 and batch size is 128, respectively. All experiments are conducted by using Pytorch framework on one NVIDIA RTX 2080Ti GPU.

\begin{table*}[]
\caption{Multimodal results of early fusion in the physiological arousal and valence evaluation on the Ulm-TSST dataset.}
\vspace{-1.0em}
\begin{tabular}{c|c|cc|cc|c}
\toprule
\multirow{2}{*}{Modality} & \multirow{2}{*}{Feature}   & \multicolumn{2}{c|}{Arousal$\uparrow$} & \multicolumn{2}{c|}{Valence$\uparrow$} & Combined$\uparrow$ \\ \cline{3-7}
  &  & Dev.  & Test  & Dev.  & Test  & Test  \\ \hline
\multirow{9}{*}{A+V} & eGeMaps+AffectNet   & 0.7800 & 0.6342 & 0.8483 & 0.6968 & 0.6655   \\
  & eGeMaps+APViT & 0.6324 & 0.4309 & 0.9203 & 0.8120 & 0.6214   \\
  & DeepSpectrum+AffectNet & 0.9253 & 0.8089 & 0.7141 & 0.5054 & 0.6571   \\
  & DeepSpectrum+APViT & 0.5714 & 0.4994 & 0.8527 & 0.6555 & 0.5775   \\
  & Wave2Vec2.0+AffectNet & 0.9093 & \textbf{0.8120} & 0.8729 & 0.6714 & 0.7417   \\
  & Wave2Vec2.0+APViT & 0.4881 & 0.3154 & 0.9168 & 0.8105 & 0.5630   \\
  & eGeMaps+AffectNet+APViT & 0.7182 & 0.4685 & 0.8076 & 0.7078 & 0.5881   \\
  & DeepSpectrum+AffectNet+APViT  & 0.7093 & 0.4842 & 0.8965 & 0.7745 & 0.6294   \\
  & Wave2Vec2.0+AffectNet+APViT & 0.7832 & 0.4745 & 0.7826 & 0.6432 & 0.5589   \\ \hline
\multirow{9}{*}{A+V+P} & eGeMaps+AffectNet+Phys & 0.7627 & 0.6138 & 0.8856 & 0.7799 & 0.6969   \\
  & eGeMaps+APViT+Phys  & 0.6466 & 0.4455 & 0.9288 & \textbf{0.8257} & 0.6313   \\
  & DeepSpectrum+AffectNet+Phys & 0.8431 & 0.5867 & 0.8024 & 0.6359 & 0.6113   \\
  & DeepSpectrum+APViT+Phys    & 0.7603 & 0.6686 & 0.8370 & 0.6838 & 0.6762   \\
  & Wave2Vec2.0+AffectNet+Phys  & 0.6323 & 0.4605 & 0.7923 & 0.5592 & 0.5059   \\
  & Wave2Vec2.0+APViT+Phys     & 0.3486 & 0.2315 & 0.9212 & 0.7574 & 0.4945   \\
  & eGeMaps+AffectNet+APViT+Phys      & 0.6773 & 0.4469 & 0.8382 & 0.7546 & 0.6008   \\
  & DeepSpectrum+AffectNet+APViT+Phys & 0.8411 & 0.6348 & 0.7926 & 0.6164 & 0.6256   \\
  & Wave2Vec2.0+AffectNet+APViT+Phys  & 0.7707 & 0.5388 & 0.7516 & 0.6160 & 0.5774   \\ \bottomrule
\end{tabular}
\vspace{-1.5em}
\label{table:t3}
\end{table*}

{\bf Evaluation Metric.} For evaluation, the concordance correlation coefficient (CCC) is used to compute the correlation between the predicted arousal or valence sequences $y_{pre}$ and the ground-truth sequences $y_{gt}$ of a video as bellow:
\begin{equation}
CCC({y_{pre},y_{gt}})=\frac{2\rho \delta_{pre}\delta_{gt}}{\delta_{pre}^{2}\delta_{gt}^{2}+(\mu_{pre}-\mu_{gt})^2},
\end{equation}
where $\mu_{pre}$ and $\mu_{gt}$ denote the mean of the prediction $y_{pre}$ and the label $y_{gt}$, respectively. $\delta_{pre}$ and $\delta_{gt}$ are their standard deviations. $\rho$ indicates the Pearson correlation coefficient (PCC) between $y_{pre}$ and $y_{gt}$. In other words, we evaluate the CCC at the video level, where $CCC \in [-1,1]$. The case of $CCC$ = 0 indicates the irrelevance between the prediction and ground truth. When $CCC$ is 1 or -1, the predicted arousal or valence value $y_{pre}$ and the ground-truth arousal or valence value $y_{gt}$ are positively or negatively correlated.

\subsection{Unimodal results}
In this section, we employ the baseline model based on RNN to perform physiological arousal and valence evaluation on features extracted from each modality.  As shown in Table~\ref{table:t2}, for the features provided by the MuSe 2023 challenge, all results are obtained according to the given parameter settings. From the experimental results, it can be observed that regardless of arousal or valence prediction, most features perform significantly lower on the test set compared to the development set.  For instance, the physiological arousal prediction based on FAU features achieves a CCC value of 0.6088 on the development set, but only 0.3476 on the test set. 

Regarding the audio modality, the CCC values of several features are higher than 0.5 on the test set, indicating the effectiveness of audio features in emotion prediction. However, in terms of video modality, some features show relatively poor emotion prediction, such as FAU, ViT, and FaceNet512, whose physiological arousal or valence values on the test data are lower than 0.5, resulting in the limitation to effectively utilize video information. In contrast, the video features we extracted, \ie, AffectNet and APVIT, achieve test results of 0.6249 and 0.7196 on arousal and valence, respectively, effectively improving the emotion estimation. Furthermore, we also validate two additional modalities, namely physiological signal and text data.  We observe severe generalization and stability issues in both physiological and text features. For example, the BERT feature performs poorly, with a CCC performance of 0.5174 on the development set, which decreases to 0.2348 on the test set, representing the worst performance among all features. Compared to the text feature, the physiological feature demonstrates better performance in both arousal and valence, with values of 0.2269 \vs 0.1613 on arousal and 0.4110 \vs 0.3102 on valence, respectively. Based on these issues, we exclude the text feature from the subsequent multimodal fusion experiments.

On the arousal dimension, the performance of the audio modality is noticeably superior to the other three modalities. These findings are consistent with the results of the baseline paper~\cite{christ2023muse}. On the valence dimension, our proposed video modality features, namely AffectNet and APViT, outperform all features significantly. Ultimately, we find that the audio feature (\ie, DeepSpectrum) and the video feature (\ie, APViT) achieve the best performance for physiological valence and arousal, with scores of 0.7447 and 0.7196, respectively. These results motivate us to explore the fusion of multimodal features to enhance the overall performance of the model in terms of both robustness and accuracy.

\subsection{Multimodal results}
In this section, we present the results of multimodal fusion for physiological arousal and valence on the development and test sets, respectively. For a fair comparison, we hold uniform hyperparameter settings, \ie, one layer RNN with a hidden size of 256. Please note that the following comparisons are based on the test set results. As shown in Table~\ref{table:t3}, we find that different multimodal features vary significantly. Some multimodal combinations outperform the unimodal features. For example, the combination of ``eGeMaps+APViT+Phys'' achieves a valence score of 0.8257, surpassing the unimodal CCC of 0.5472, 0.7196, and 0.4110. However, several fusion features perform even worse than unimodal features. For instance, the combination of ``Wave2Vec2.0+APViT'' achieves arousal of only 0.3154, lower than the unimodal of 0.5053 and 0.5413. In addition, we also find that stacking multiple video modality features does not always yield better results. For example, the multimodal combination of ``DeepSpectrum+AffectNet+APViT'' (arousal of 0.4842) performs worse than ``DeepSpectrum+AffectNet'' (arousal of 0.8089) and ``DeepSpectrum+APViT'' (arousal of 0.4994). As expected, physiological signals can improve the accuracy of emotion prediction in certain multimodal fusions. Compared to ``DeepSpectrum+APViT'', the arousal is increased by 0.1689 after fusing the Phys feature.

From the experimental results in Table~\ref{table:t3}, we obtain the best performance of 0.8120 for arousal prediction by fusing audio features (Wave2Vec2.0) and video features (AffectNet).  This demonstrates that these two features can achieve further performance improvement by integrating complementary emotion-related information.  Furthermore, APVIT displays poor performance in arousal prediction which is consistent with the unimodal results in Table 3.  Such as fusing eGeMaps and APVIT features, its performance is inferior to the unimodal (0.4309 <  0.5523, 0.5413).  In contrast, AffectNet features facilitate arousal prediction, such as the multimodal fusion of ``Wave2Vec2.0+AffectNet'' and ``DeepSpectrum+AffectNet''.  Additionally, in most cases, physiological signals are conducive to arousal prediction and can even hinder multimodal fusion performance.   Regard of valence, the fusion of ``eGeMaps+APViT+Phys'' achieves superior performance with a CCC of 0.8257.  Compared to ``eGeMaps+APViT,'' the additional physiological signals improve performance by 0.0137, demonstrating a higher correlation between physiological signals and valence.  In summary, AffectNet is more suitable for arousal prediction, while APVIT is more beneficial for valence prediction.  By appropriately fusing multimodal features, we can achieve improved performance in emotion estimation.

\begin{table}[]
\caption{Ablation study for hyper-parameter setting in layer, model\_dim, and rnn\_bi on the arousal dimension estimation. The evaluation is performed with the wave2vec2.0 and AffectNet features. The rnn\_bi denotes whether the RNN is bidirectional, ``Y''denotes bidirectional RNN, ``N'' denotes unidirectional RNN.}
\vspace{-1.0em}
\begin{tabular}{c|c|c|cc}
\toprule
\multirow{2}{*}{\#Layers} & \multirow{2}{*}{model\_dim} & \multirow{2}{*}{rnn\_bi} & \multicolumn{2}{c}{Arousal$\uparrow$} \\ \cline{4-5}
  &   &  & Dev.  & Test  \\ \hline
\multirow{4}{*}{1} & \multirow{2}{*}{128} & N   & 0.7468 & 0.6089 \\
  &   & Y   & 0.6888 & 0.4197 \\ \cline{2-5} 
  & \multirow{2}{*}{256} & N   & 0.9093 & 0.8120 \\
  &   & Y   & 0.7356 & 0.5569 \\ \hline
\multirow{4}{*}{2} & \multirow{2}{*}{128} & N   & 0.8248 & 0.6009 \\
  &   & Y   & 0.7849 & 0.6059 \\ \cline{2-5} 
  & \multirow{2}{*}{256} & N   & 0.8487 & 0.6911 \\
  &   & Y   & 0.6483 & 0.5175 \\ \bottomrule
\end{tabular}
\vspace{-1.5em}
\label{table:t4}
\end{table}

\begin{table}[]
\centering
\caption{Ablation study for hyper-parameter setting in layer, model\_dim, and on the valence dimension estimation. The evaluation is performed with the eGeMaps, APViT, and Phys features. The rnn\_bi denotes whether the RNN is bidirectional, ``Y''denotes bidirectional RNN, ``N'' denotes standard RNN.}
\vspace{-1.0em}
\begin{tabular}{c|c|c|cc}
\toprule
\multirow{2}{*}{\#Layers} & \multirow{2}{*}{model\_dim} & \multirow{2}{*}{rnn\_bi} & \multicolumn{2}{c}{Valence$\uparrow$} \\ \cline{4-5}
  &   &  & Dev.  & Test  \\ \hline
\multirow{4}{*}{1} & \multirow{2}{*}{128} & N   & 0.7800 & 0.6280 \\
  &   & Y   & 0.8883 & 0.7895 \\ \cline{2-5} 
  & \multirow{2}{*}{256} & N   & 0.9023 & 0.8086 \\
  &   & Y   & 0.9106 & 0.8172 \\ \hline
\multirow{4}{*}{2} & \multirow{2}{*}{128} & N   & 0.4735 & 0.4338 \\
  &   & Y   & 0.8785 & 0.7631 \\ \cline{2-5} 
  & \multirow{2}{*}{256} & N   & 0.7151 & 0.6176 \\
  &   & Y   & 0.8800 & 0.8007 \\ \bottomrule
\end{tabular}
\vspace{-1.5em}
\label{table:t5}
\end{table}

\begin{table}[]
\centering
\caption{The experimental results on the test set of the Ulm-TSST dataset. The results of Top-1 and Top-2 are from the Codalab competition page.}
\vspace{-1.0em}
\begin{tabular}{c|c|ccc}
\toprule
Method & Rank & Arousal & Valence & Combined\\ \hline
Baseline  & - & 0.7482 & 0.7827 & 0.7654 \\ \hline
Anonymous  & 1 & 0.86 & 0.88 & 0.87 \\
Anonymous  & 2 & 0.83 & 0.89 & 0.86\\ \hline
\textbf{Ensemble (Ours)}  & 3 & 0.8386 & 0.8492 & 0.8439 \\ \bottomrule
\end{tabular}
\vspace{-1.5em}
\label{table:t6}
\end{table}

\subsection{Ablation Study}
To evaluate the impact and performance of hyper-parameter settings, we conduct ablation studies in this section. There are three important hyper-parameters for the network: the layer of RNN, the dimension of model, and whether is bidirectional RNN. As shown in Table~\ref{table:t4}, \ref{table:t5}, we divide the configurations into 4 groups row-wise, and different values are assigned for one hyper-parameters while keeping the other two hyper-parameters fixed to evaluate the impact and choice of each configuration. As illustrated in Table~\ref{table:t4}, our model could achieve higher arousal when Layer=1, model\_dim=256, and standard RNN is selected. Surprisingly, we find that bidirectional RNNs suppress the model's predictions for arousal. In addition, the number of layers is not directly proportional to the performance, and merely 1 is the most appropriate value. The high number of layers will easily cause the model to overfit. Compared with the number of layers, a larger model dimension can represent richer emotional information, which is worthwhile for arousal prediction. Regarding the ablation experiment of valence, Table~\ref{table:t5} shows that the model achieves the best performance with layer 1 and model\_dim 256, which is similar to the arousal prediction provided in Table~\ref{table:t4}. Besides, the bidirectional RNN can further improve the prediction of valence (0.8086 to 0.8172).

\subsection{Submission Results}
Finally, Table~\ref{table:t6} shows the test results of other teams in the Valence-Arousal estimation of the MuSe-Personalisation sub-challenge, and our proposed solution achieves the 3rd place performance. For our final submission results, we apply an ensemble strategy with several feature combinations to improve the performance. Specifically, for arousal prediction, we ensemble two feature combinations, \ie, ``DeepSpectrum+AffectNet'' and ``Wave2Vec2.0+AffectNet'', which improved by 0.0904 (from 0.7482 to 0.8386), compared to baseline results. For valence prediction, we ensemble three feature combinations, \ie, ``eGeMaps+APViT'', ``Wave2Vec2.0+APV'', and ``eGeMaps+APViT+Phys'', which improved by 0.0665 (from 0.7827 to 0.8492), compared to baseline results. Compared to all other teams in all parameters, we achieved 2nd place performance in arousal evaluation and 4th place performance in valence.

\section{Conclusions}~\label{sec:conclusion}
In this paper, we present our solution developed for the MuSe-Personalisation sub-challenge in MuSe 2023. Our method utilizes multimodal information (\ie, audio, video, and physiological signals) and early fusion approach to capture emotion-related personalisation. Furthermore, a model ensemble strategy is used to improve the predictions. The experiment results show that our solution achieves \textbf{0.8386} (rank \textbf{Top 2}) CCC for arousal, \textbf{0.8492} (rank \textbf{Top 4}) CCC for valence, and \textbf{0.8439} (rank \textbf{Top 3}) CCC for their mean on the test set of Ulm-TSST dataset, outperforming the baseline system by a large margin (\ie, 0.7482, 0.7827, and 0.7654).
In the future, we plan to address the MuSe-Personalisation from other perspectives, \eg, new physiological feature~\cite{qian2023dual}, skeleton-based method~\cite{li2023joint} for action modeling, data augmentation for imbalanced data distribution, and end-to-end personalisation prediction without any feature pre-extraction.  

\begin{acks}
This work was supported in part by the National Key R\&D Program of China (2022YFB4500600), in part by the National Natural Science Foundation of China (62202139, 62272144, 72188101, 62020106007, and U20A20183), in part by the Anhui Provincial Natural Science Foundation (2208085QF191), and in part by the Major Project of Anhui Province (202203a05020011). 
\end{acks}

\bibliographystyle{ACM-Reference-Format}
\balance
\bibliography{sample-base}


\begin{thebibliography}{29}


\ifx \showCODEN    \undefined \def \showCODEN     #1{\unskip}     \fi
\ifx \showDOI      \undefined \def \showDOI       #1{#1}\fi
\ifx \showISBNx    \undefined \def \showISBNx     #1{\unskip}     \fi
\ifx \showISBNxiii \undefined \def \showISBNxiii  #1{\unskip}     \fi
\ifx \showISSN     \undefined \def \showISSN      #1{\unskip}     \fi
\ifx \showLCCN     \undefined \def \showLCCN      #1{\unskip}     \fi
\ifx \shownote     \undefined \def \shownote      #1{#1}          \fi
\ifx \showarticletitle \undefined \def \showarticletitle #1{#1}   \fi
\ifx \showURL      \undefined \def \showURL       {\relax}        \fi
\providecommand\bibfield[2]{#2}
\providecommand\bibinfo[2]{#2}
\providecommand\natexlab[1]{#1}
\providecommand\showeprint[2][]{arXiv:#2}

\bibitem[Amiriparian et~al\mbox{.}(2022)]%
        {amiriparian2022muse}
\bibfield{author}{\bibinfo{person}{Shahin Amiriparian}, \bibinfo{person}{Lukas
  Christ}, \bibinfo{person}{Andreas K{\"o}nig}, \bibinfo{person}{Eva-Maria
  Me{\ss}ner}, \bibinfo{person}{Alan Cowen}, \bibinfo{person}{Erik Cambria},
  {and} \bibinfo{person}{Bj{\"o}rn~W Schuller}.}
  \bibinfo{year}{2022}\natexlab{}.
\newblock \showarticletitle{Muse 2022 challenge: Multimodal humour, emotional
  reactions, and stress}. In \bibinfo{booktitle}{\emph{Proceedings of the 30th
  ACM International Conference on Multimedia}}. \bibinfo{pages}{7389--7391}.
\newblock


\bibitem[Amiriparian et~al\mbox{.}(2017)]%
        {amiriparian2017snore}
\bibfield{author}{\bibinfo{person}{Shahin Amiriparian},
  \bibinfo{person}{Maurice Gerczuk}, \bibinfo{person}{Sandra Ottl},
  \bibinfo{person}{Nicholas Cummins}, \bibinfo{person}{Michael Freitag},
  \bibinfo{person}{Sergey Pugachevskiy}, \bibinfo{person}{Alice Baird}, {and}
  \bibinfo{person}{Bj{\"o}rn Schuller}.} \bibinfo{year}{2017}\natexlab{}.
\newblock \showarticletitle{Snore sound classification using image-based deep
  spectrum features}.
\newblock  (\bibinfo{year}{2017}).
\newblock


\bibitem[Ayata et~al\mbox{.}(2020)]%
        {ayata2020emotion}
\bibfield{author}{\bibinfo{person}{De{\u{g}}er Ayata}, \bibinfo{person}{Yusuf
  Yaslan}, {and} \bibinfo{person}{Mustafa~E Kamasak}.}
  \bibinfo{year}{2020}\natexlab{}.
\newblock \showarticletitle{Emotion recognition from multimodal physiological
  signals for emotion aware healthcare systems}.
\newblock \bibinfo{journal}{\emph{Journal of Medical and Biological
  Engineering}}  \bibinfo{volume}{40} (\bibinfo{year}{2020}),
  \bibinfo{pages}{149--157}.
\newblock


\bibitem[Baevski et~al\mbox{.}(2020)]%
        {baevski2020wav2vec}
\bibfield{author}{\bibinfo{person}{Alexei Baevski}, \bibinfo{person}{Yuhao
  Zhou}, \bibinfo{person}{Abdelrahman Mohamed}, {and} \bibinfo{person}{Michael
  Auli}.} \bibinfo{year}{2020}\natexlab{}.
\newblock \showarticletitle{wav2vec 2.0: A framework for self-supervised
  learning of speech representations}.
\newblock \bibinfo{journal}{\emph{Advances in neural information processing
  systems}}  \bibinfo{volume}{33} (\bibinfo{year}{2020}),
  \bibinfo{pages}{12449--12460}.
\newblock


\bibitem[Caron et~al\mbox{.}(2021)]%
        {caron2021vit}
\bibfield{author}{\bibinfo{person}{Mathilde Caron}, \bibinfo{person}{Hugo
  Touvron}, \bibinfo{person}{Ishan Misra}, \bibinfo{person}{Hervé Jegou},
  \bibinfo{person}{Julien Mairal}, \bibinfo{person}{Piotr Bojanowski}, {and}
  \bibinfo{person}{Armand Joulin}.} \bibinfo{year}{2021}\natexlab{}.
\newblock \showarticletitle{Emerging Properties in Self-Supervised Vision
  Transformers}. In \bibinfo{booktitle}{\emph{2021 IEEE/CVF International
  Conference on Computer Vision (ICCV)}}. \bibinfo{pages}{9630--9640}.
\newblock


\bibitem[Christ et~al\mbox{.}(2023)]%
        {christ2023muse}
\bibfield{author}{\bibinfo{person}{Lukas Christ}, \bibinfo{person}{Shahin
  Amiriparian}, \bibinfo{person}{Alice Baird}, \bibinfo{person}{Alexander
  Kathan}, \bibinfo{person}{Niklas M{\"u}ller}, \bibinfo{person}{Steffen Klug},
  \bibinfo{person}{Chris Gagne}, \bibinfo{person}{Panagiotis Tzirakis},
  \bibinfo{person}{Eva-Maria Me{\ss}ner}, \bibinfo{person}{Andreas K{\"o}nig},
  {et~al\mbox{.}}} \bibinfo{year}{2023}\natexlab{}.
\newblock \showarticletitle{The MuSe 2023 Multimodal Sentiment Analysis
  Challenge: Mimicked Emotions, Cross-Cultural Humour, and Personalisation}.
\newblock \bibinfo{journal}{\emph{arXiv preprint arXiv:2305.03369}}
  (\bibinfo{year}{2023}).
\newblock


\bibitem[Christ et~al\mbox{.}(2022)]%
        {christ2022muse}
\bibfield{author}{\bibinfo{person}{Lukas Christ}, \bibinfo{person}{Shahin
  Amiriparian}, \bibinfo{person}{Alice Baird}, \bibinfo{person}{Panagiotis
  Tzirakis}, \bibinfo{person}{Alexander Kathan}, \bibinfo{person}{Niklas
  M{\"u}ller}, \bibinfo{person}{Lukas Stappen}, \bibinfo{person}{Eva-Maria
  Me{\ss}ner}, \bibinfo{person}{Andreas K{\"o}nig}, \bibinfo{person}{Alan
  Cowen}, {et~al\mbox{.}}} \bibinfo{year}{2022}\natexlab{}.
\newblock \showarticletitle{The muse 2022 multimodal sentiment analysis
  challenge: humor, emotional reactions, and stress}. In
  \bibinfo{booktitle}{\emph{Proceedings of the 3rd International on Multimodal
  Sentiment Analysis Workshop and Challenge}}. \bibinfo{pages}{5--14}.
\newblock


\bibitem[Ekman and Friesen(1978)]%
        {ekman1978facial}
\bibfield{author}{\bibinfo{person}{Paul Ekman} {and} \bibinfo{person}{Wallace~V
  Friesen}.} \bibinfo{year}{1978}\natexlab{}.
\newblock \showarticletitle{Facial action coding system}.
\newblock \bibinfo{journal}{\emph{Environmental Psychology \& Nonverbal
  Behavior}} (\bibinfo{year}{1978}).
\newblock


\bibitem[Eyben et~al\mbox{.}(2010)]%
        {eyben2010toolkit}
\bibfield{author}{\bibinfo{person}{Florian Eyben}, \bibinfo{person}{Martin
  W\"{o}llmer}, {and} \bibinfo{person}{Bj\"{o}rn Schuller}.}
  \bibinfo{year}{2010}\natexlab{}.
\newblock \showarticletitle{Opensmile: The Munich Versatile and Fast
  Open-Source Audio Feature Extractor}. In
  \bibinfo{booktitle}{\emph{Proceedings of the 18th ACM International
  Conference on Multimedia}}. \bibinfo{publisher}{Association for Computing
  Machinery}, \bibinfo{address}{New York, NY, USA},
  \bibinfo{pages}{1459–1462}.
\newblock
\showISBNx{9781605589336}


\bibitem[Guo et~al\mbox{.}(2018)]%
        {guo2018hierarchical}
\bibfield{author}{\bibinfo{person}{Dan Guo}, \bibinfo{person}{Wengang Zhou},
  \bibinfo{person}{Houqiang Li}, {and} \bibinfo{person}{Meng Wang}.}
  \bibinfo{year}{2018}\natexlab{}.
\newblock \showarticletitle{Hierarchical LSTM for sign language translation}.
  In \bibinfo{booktitle}{\emph{Proceedings of the AAAI conference on artificial
  intelligence}}, Vol.~\bibinfo{volume}{32}.
\newblock


\bibitem[Huang et~al\mbox{.}(2017)]%
        {huang2017densely}
\bibfield{author}{\bibinfo{person}{Gao Huang}, \bibinfo{person}{Zhuang Liu},
  \bibinfo{person}{Laurens Van Der~Maaten}, {and} \bibinfo{person}{Kilian~Q
  Weinberger}.} \bibinfo{year}{2017}\natexlab{}.
\newblock \showarticletitle{Densely connected convolutional networks}. In
  \bibinfo{booktitle}{\emph{Proceedings of the IEEE Conference on Computer
  Vision and Pattern Recognition}}. \bibinfo{pages}{4700--4708}.
\newblock


\bibitem[Li et~al\mbox{.}(2023a)]%
        {li2023multimodal}
\bibfield{author}{\bibinfo{person}{Jia Li}, \bibinfo{person}{Yin Chen},
  \bibinfo{person}{Xuesong Zhang}, \bibinfo{person}{Jiantao Nie},
  \bibinfo{person}{Ziqiang Li}, \bibinfo{person}{Yangchen Yu},
  \bibinfo{person}{Yan Zhang}, \bibinfo{person}{Richang Hong}, {and}
  \bibinfo{person}{Meng Wang}.} \bibinfo{year}{2023}\natexlab{a}.
\newblock \showarticletitle{Multimodal feature extraction and fusion for
  emotional reaction intensity estimation and expression classification in
  videos with transformers}. In \bibinfo{booktitle}{\emph{Proceedings of the
  IEEE/CVF Conference on Computer Vision and Pattern Recognition}}.
  \bibinfo{pages}{5837--5843}.
\newblock


\bibitem[Li et~al\mbox{.}(2022)]%
        {li2022hybrid}
\bibfield{author}{\bibinfo{person}{Jia Li}, \bibinfo{person}{Ziyang Zhang},
  \bibinfo{person}{Junjie Lang}, \bibinfo{person}{Yueqi Jiang},
  \bibinfo{person}{Liuwei An}, \bibinfo{person}{Peng Zou},
  \bibinfo{person}{Yangyang Xu}, \bibinfo{person}{Sheng Gao},
  \bibinfo{person}{Jie Lin}, \bibinfo{person}{Chunxiao Fan}, {et~al\mbox{.}}}
  \bibinfo{year}{2022}\natexlab{}.
\newblock \showarticletitle{Hybrid multimodal feature extraction, mining and
  fusion for sentiment analysis}. In \bibinfo{booktitle}{\emph{Proceedings of
  the 3rd International on Multimodal Sentiment Analysis Workshop and
  Challenge}}. \bibinfo{pages}{81--88}.
\newblock


\bibitem[Li et~al\mbox{.}(2023c)]%
        {li2023joint}
\bibfield{author}{\bibinfo{person}{Kun Li}, \bibinfo{person}{Dan Guo},
  \bibinfo{person}{Guoliang Chen}, \bibinfo{person}{Xinge Peng}, {and}
  \bibinfo{person}{Meng Wang}.} \bibinfo{year}{2023}\natexlab{c}.
\newblock \showarticletitle{Joint Skeletal and Semantic Embedding Loss for
  Micro-gesture Classification}.
\newblock \bibinfo{journal}{\emph{arXiv preprint arXiv:2307.10624}}
  (\bibinfo{year}{2023}).
\newblock


\bibitem[Li et~al\mbox{.}(2021)]%
        {li2021proposal}
\bibfield{author}{\bibinfo{person}{Kun Li}, \bibinfo{person}{Dan Guo}, {and}
  \bibinfo{person}{Meng Wang}.} \bibinfo{year}{2021}\natexlab{}.
\newblock \showarticletitle{Proposal-free video grounding with contextual
  pyramid network}. In \bibinfo{booktitle}{\emph{Proceedings of the AAAI
  Conference on Artificial Intelligence}}, Vol.~\bibinfo{volume}{35}.
  \bibinfo{pages}{1902--1910}.
\newblock


\bibitem[Li et~al\mbox{.}(2023b)]%
        {li2023vigt}
\bibfield{author}{\bibinfo{person}{Kun Li}, \bibinfo{person}{Dan Guo}, {and}
  \bibinfo{person}{Meng Wang}.} \bibinfo{year}{2023}\natexlab{b}.
\newblock \showarticletitle{ViGT: Proposal-free Video Grounding with Learnable
  Token in Transformer}.
\newblock \bibinfo{journal}{\emph{arXiv preprint arXiv:2308.06009}}
  (\bibinfo{year}{2023}).
\newblock


\bibitem[Li et~al\mbox{.}(2017)]%
        {li2017reliable}
\bibfield{author}{\bibinfo{person}{Shan Li}, \bibinfo{person}{Weihong Deng},
  {and} \bibinfo{person}{JunPing Du}.} \bibinfo{year}{2017}\natexlab{}.
\newblock \showarticletitle{Reliable crowdsourcing and deep locality-preserving
  learning for expression recognition in the wild}. In
  \bibinfo{booktitle}{\emph{Proceedings of the IEEE conference on computer
  vision and pattern recognition}}. \bibinfo{pages}{2852--2861}.
\newblock


\bibitem[Makowski et~al\mbox{.}(2021)]%
        {makowski2021neurokit2}
\bibfield{author}{\bibinfo{person}{Dominique Makowski}, \bibinfo{person}{Tam
  Pham}, \bibinfo{person}{Zen~J Lau}, \bibinfo{person}{Jan~C Brammer},
  \bibinfo{person}{Fran{\c{c}}ois Lespinasse}, \bibinfo{person}{Hung Pham},
  \bibinfo{person}{Christopher Sch{\"o}lzel}, {and} \bibinfo{person}{SH~Annabel
  Chen}.} \bibinfo{year}{2021}\natexlab{}.
\newblock \showarticletitle{NeuroKit2: A Python toolbox for neurophysiological
  signal processing}.
\newblock \bibinfo{journal}{\emph{Behavior research methods}}
  (\bibinfo{year}{2021}), \bibinfo{pages}{1--8}.
\newblock


\bibitem[Mao et~al\mbox{.}(2023)]%
        {mao2023poster}
\bibfield{author}{\bibinfo{person}{Jiawei Mao}, \bibinfo{person}{Rui Xu},
  \bibinfo{person}{Xuesong Yin}, \bibinfo{person}{Yuanqi Chang},
  \bibinfo{person}{Binling Nie}, {and} \bibinfo{person}{Aibin Huang}.}
  \bibinfo{year}{2023}\natexlab{}.
\newblock \bibinfo{title}{POSTER++: A simpler and stronger facial expression
  recognition network}.
\newblock
\newblock
\showeprint[arxiv]{2301.12149}~[cs.CV]


\bibitem[Mollahosseini et~al\mbox{.}(2017)]%
        {mollahosseini2017affectnet}
\bibfield{author}{\bibinfo{person}{Ali Mollahosseini}, \bibinfo{person}{Behzad
  Hasani}, {and} \bibinfo{person}{Mohammad~H Mahoor}.}
  \bibinfo{year}{2017}\natexlab{}.
\newblock \showarticletitle{Affectnet: A database for facial expression,
  valence, and arousal computing in the wild}.
\newblock \bibinfo{journal}{\emph{IEEE Transactions on Affective Computing}}
  \bibinfo{volume}{10}, \bibinfo{number}{1} (\bibinfo{year}{2017}),
  \bibinfo{pages}{18--31}.
\newblock


\bibitem[Qian et~al\mbox{.}(2023)]%
        {qian2023dual}
\bibfield{author}{\bibinfo{person}{Wei Qian}, \bibinfo{person}{Dan Guo},
  \bibinfo{person}{Kun Li}, \bibinfo{person}{Xilan Tian}, {and}
  \bibinfo{person}{Meng Wang}.} \bibinfo{year}{2023}\natexlab{}.
\newblock \showarticletitle{Dual-path TokenLearner for Remote
  Photoplethysmography-based Physiological Measurement with Facial Videos}.
\newblock \bibinfo{journal}{\emph{arXiv preprint arXiv:2308.07771}}
  (\bibinfo{year}{2023}).
\newblock


\bibitem[Santamaria-Granados et~al\mbox{.}(2018)]%
        {santamaria2018using}
\bibfield{author}{\bibinfo{person}{Luz Santamaria-Granados},
  \bibinfo{person}{Mario Munoz-Organero}, \bibinfo{person}{Gustavo
  Ramirez-Gonzalez}, \bibinfo{person}{Enas Abdulhay}, {and}
  \bibinfo{person}{NJIA Arunkumar}.} \bibinfo{year}{2018}\natexlab{}.
\newblock \showarticletitle{Using deep convolutional neural network for emotion
  detection on a physiological signals dataset (AMIGOS)}.
\newblock \bibinfo{journal}{\emph{IEEE Access}}  \bibinfo{volume}{7}
  (\bibinfo{year}{2018}), \bibinfo{pages}{57--67}.
\newblock


\bibitem[Schroff et~al\mbox{.}(2015)]%
        {schroff2015facenet}
\bibfield{author}{\bibinfo{person}{Florian Schroff}, \bibinfo{person}{Dmitry
  Kalenichenko}, {and} \bibinfo{person}{James Philbin}.}
  \bibinfo{year}{2015}\natexlab{}.
\newblock \showarticletitle{FaceNet: A unified embedding for face recognition
  and clustering}. In \bibinfo{booktitle}{\emph{2015 IEEE Conference on
  Computer Vision and Pattern Recognition}}. \bibinfo{pages}{815--823}.
\newblock


\bibitem[Stappen et~al\mbox{.}(2021)]%
        {stappen2021muse}
\bibfield{author}{\bibinfo{person}{Lukas Stappen}, \bibinfo{person}{Alice
  Baird}, \bibinfo{person}{Lukas Christ}, \bibinfo{person}{Lea Schumann},
  \bibinfo{person}{Benjamin Sertolli}, \bibinfo{person}{Eva-Maria Messner},
  \bibinfo{person}{Erik Cambria}, \bibinfo{person}{Guoying Zhao}, {and}
  \bibinfo{person}{Bj{\"o}rn~W Schuller}.} \bibinfo{year}{2021}\natexlab{}.
\newblock \showarticletitle{The MuSe 2021 multimodal sentiment analysis
  challenge: sentiment, emotion, physiological-emotion, and stress}.
\newblock In \bibinfo{booktitle}{\emph{Proceedings of the 2nd on Multimodal
  Sentiment Analysis Challenge}}. \bibinfo{pages}{5--14}.
\newblock


\bibitem[Valstar et~al\mbox{.}(2016)]%
        {valstar2016avec}
\bibfield{author}{\bibinfo{person}{Michel Valstar}, \bibinfo{person}{Jonathan
  Gratch}, \bibinfo{person}{Bj{\"o}rn Schuller}, \bibinfo{person}{Fabien
  Ringeval}, \bibinfo{person}{Denis Lalanne}, \bibinfo{person}{Mercedes
  Torres~Torres}, \bibinfo{person}{Stefan Scherer}, \bibinfo{person}{Giota
  Stratou}, \bibinfo{person}{Roddy Cowie}, {and} \bibinfo{person}{Maja
  Pantic}.} \bibinfo{year}{2016}\natexlab{}.
\newblock \showarticletitle{Avec 2016: Depression, mood, and emotion
  recognition workshop and challenge}. In \bibinfo{booktitle}{\emph{Proceedings
  of the 6th international workshop on audio/visual emotion challenge}}.
  \bibinfo{pages}{3--10}.
\newblock


\bibitem[Xue et~al\mbox{.}(2021)]%
        {xue2021transfer}
\bibfield{author}{\bibinfo{person}{Fanglei Xue}, \bibinfo{person}{Qiangchang
  Wang}, {and} \bibinfo{person}{Guodong Guo}.} \bibinfo{year}{2021}\natexlab{}.
\newblock \showarticletitle{Transfer: Learning relation-aware facial expression
  representations with transformers}. In \bibinfo{booktitle}{\emph{Proceedings
  of the IEEE/CVF International Conference on Computer Vision}}.
  \bibinfo{pages}{3601--3610}.
\newblock


\bibitem[Xue et~al\mbox{.}(2022a)]%
        {xue2022vision}
\bibfield{author}{\bibinfo{person}{Fanglei Xue}, \bibinfo{person}{Qiangchang
  Wang}, \bibinfo{person}{Zichang Tan}, \bibinfo{person}{Zhongsong Ma}, {and}
  \bibinfo{person}{Guodong Guo}.} \bibinfo{year}{2022}\natexlab{a}.
\newblock \showarticletitle{Vision transformer with attentive pooling for
  robust facial expression recognition}.
\newblock \bibinfo{journal}{\emph{IEEE Transactions on Affective Computing}}
  (\bibinfo{year}{2022}).
\newblock


\bibitem[Xue et~al\mbox{.}(2022b)]%
        {apvit}
\bibfield{author}{\bibinfo{person}{Fanglei Xue}, \bibinfo{person}{Qiangchang
  Wang}, \bibinfo{person}{Zichang Tan}, \bibinfo{person}{Zhongsong Ma}, {and}
  \bibinfo{person}{Guodong Guo}.} \bibinfo{year}{2022}\natexlab{b}.
\newblock \showarticletitle{Vision Transformer with Attentive Pooling for
  Robust Facial Expression Recognition}.
\newblock \bibinfo{journal}{\emph{IEEE Transactions on Affective Computing}}
  (\bibinfo{year}{2022}).
\newblock


\bibitem[Zitouni et~al\mbox{.}(2021)]%
        {zitouni2021arousal}
\bibfield{author}{\bibinfo{person}{M~Sami Zitouni},
  \bibinfo{person}{Cheul~Young Park}, \bibinfo{person}{Uichin Lee},
  \bibinfo{person}{Leontios Hadjileontiadis}, {and} \bibinfo{person}{Ahsan
  Khandoker}.} \bibinfo{year}{2021}\natexlab{}.
\newblock \showarticletitle{Arousal-valence classification from peripheral
  physiological signals using long short-term memory networks}. In
  \bibinfo{booktitle}{\emph{2021 43rd Annual International Conference of the
  IEEE Engineering in Medicine \& Biology Society (EMBC)}}. IEEE,
  \bibinfo{pages}{686--689}.
\newblock


\end{thebibliography}

\end{document}